\documentclass{article}
\usepackage{spconf,amsmath,epsfig}
\usepackage{multirow}
\usepackage{cite}
\usepackage{setspace}

\usepackage{hyperref}

\begin{document}\sloppy

\def\x{{\mathbf x}}
\def\L{{\cal L}}

%
\title{EMface: Detecting Hard Faces by Exploring Receptive Field Pyraminds}
%
%

\name{Leilei Cao, Yao Xiao, and  Lin Xu$^{*}$\thanks{* Contact Author}\thanks{L. Cao, Y. Xiao, and L. Xu are with the Institute of Artificial Intelligence, Shanghai Em-Data Technology Co., Ltd., Shanghai 200000, China. E-mail: \{leilei.cao01, xiaoyao.xyz, lin.xu5470\}@gmail.com.}
}
\address{}
\maketitle
%

%
%

\begin{abstract}
Scale variation is one of the most challenging problems in face detection. Modern face detectors employ feature pyramids to deal with scale variation. However,  it might break the feature consistency across different scales of faces. In this paper, we propose a simple yet effective method named receptive field pyramids (RFP) method to enhance the representation ability of feature pyramids. It can learn different receptive fields in each feature map adaptively based on the varying scales of detected faces.
Empirical results on two face detection benchmark datasets, i.e., WIDER FACE and UFDD, demonstrate that our proposed method can accelerate the inference rate significantly while achieving state-of-the-art performance. The source code of our method is available at 
\url{https://github.com/emdata-ailab/EMface}.
\end{abstract}

\section{Introduction}
Face detection is a highly competitive research topic in computer vision \cite{Wu2018, CaoK2018, Liu2017}. In the recent, with the remarkable progress in deep convolutional neural networks (CNNs) \cite{Simonyan14, He2016, Hu_2018_CVPR} and the creation of large-scale annotated datasets \cite{imagenet_cvpr09, COCO, Yang2016}, many advanced methods\cite{Liu2016, Lin2017, dai17dcn, Liu_2018_ECCV, ZhuC2019} have been developed for visual recognition tasks. The joint learned deep feature representation and semantical metric yield significant improvements in the community of face detection \cite{ZhangK2016, Zhang_ICCV2017, Najibi_2017_ICCV, Tang_2018_ECCV, DSFD2019}. 
\begin{figure}[htbp]
\centering
\includegraphics[width=\linewidth]{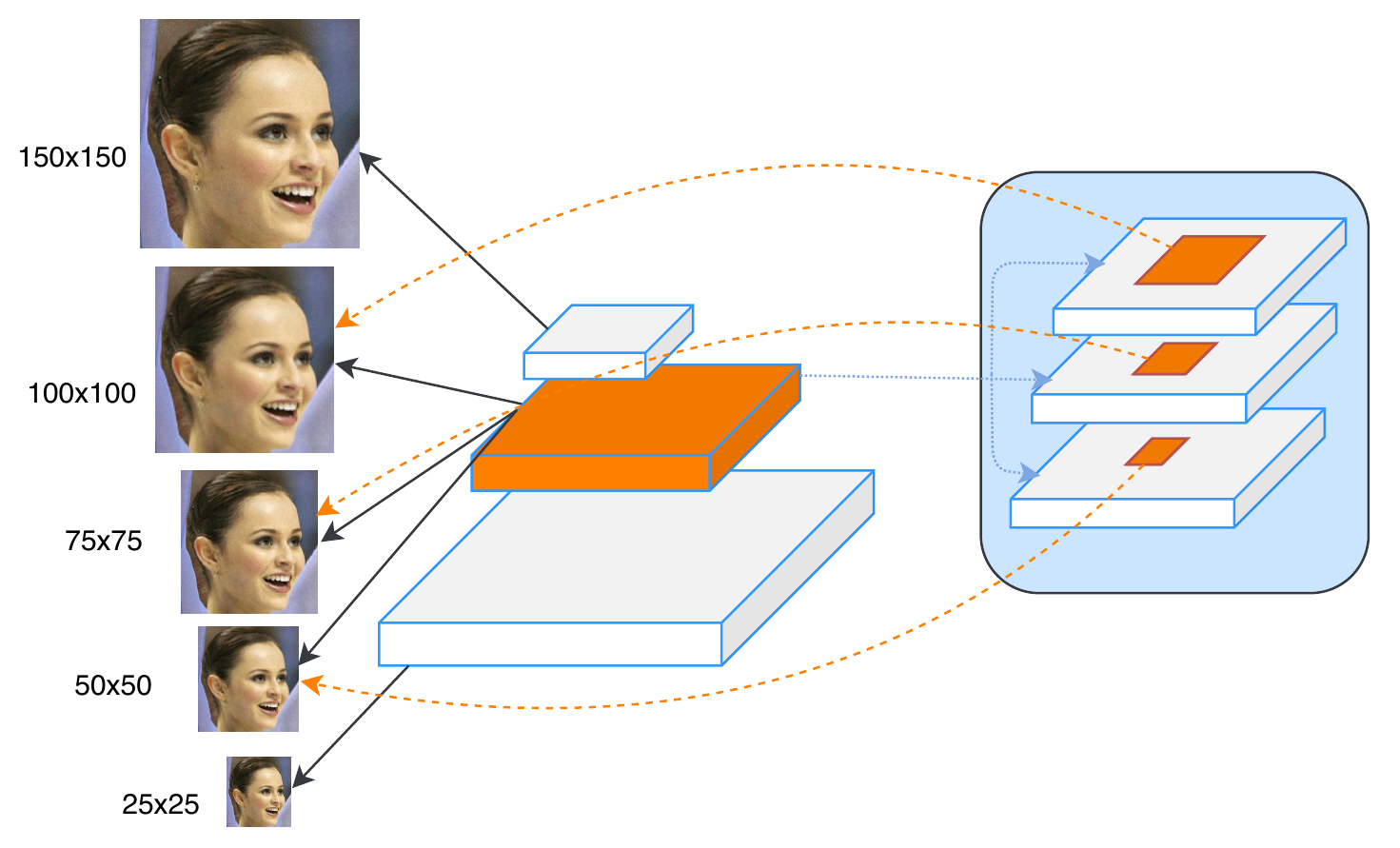}
\caption{Faces with varying scales are caught by different layers in feature pyramids, e.g.,  faces with size 50$\times$50,  75$\times$75, and 100$\times$100 pixels might be assigned to the same middle-level feature map.  It would affect the robustness of the feature pyramids to the scale variation.  We propose a receptive field pyramids (RFP) method in each feature map to enhance the representation ability of feature pyramids.}
\label{motivation}
\end{figure}

One of the most challenging problems in face detection is the effect of scale variation \cite{WangJ2018}. The early-stage works employed the image pyramids method to relieve scale variation \cite{Li2015, ZhangK2016}. The multi-scale input images in image pyramids cause a considerable increase of inference time \cite{ZhangK2016}, which makes it infeasible for practical applications. Modern face detectors employ feature pyramids to deal with scale variation \cite{Zhang_ICCV2017, Najibi_2017_ICCV, Tang_2018_ECCV, DSFD2019}. They use multi-scale feature maps from different CNN layers to detect faces with different scales. The representation ability of feature pyramids is not as powerful as image pyramids. 
One of the crucial reasons is that feature pyramids may break the consistency across different scales. Faces with various scales are assigned to a feature map based on the Intersection-over-Union (IoU) overlapping with anchor boxes. It could discretize the continuous space of possible face bounding boxes into a finite number of boxes with pre-defined locations and shapes \cite{ZhuC2019}. As Figure \ref{motivation} shown, larger faces are typically detected in high-level feature maps, while smaller faces are caught in low-level ones. Faces with a specific range of scales (e.g., faces with 50$\times$50, 75$\times$75, and 100$\times$100 pixels) could often fall into a same-level feature map.  It would substantially affect the robustness of the feature pyramids to the scale variation. 
\begin{figure}[htbp]
\centering
\includegraphics[width=\linewidth]{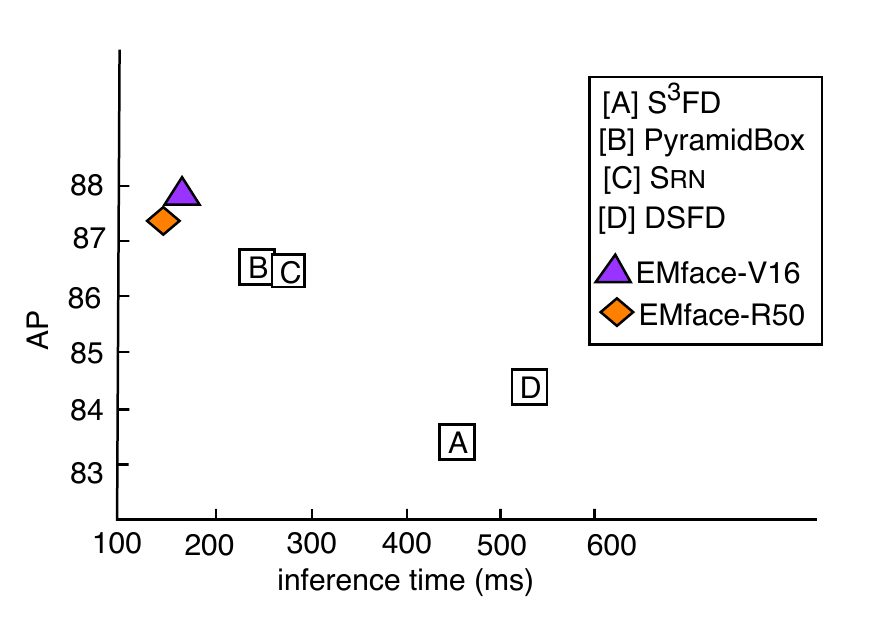}
\caption{Accuracy versus speed on the WIDER FACE validation hard subset. Our proposed face detector outperforms the state-of-the-art face detectors on the \textbf{\textit{single-scale}} testing in terms of acuracy and speed. Details are shown in Table \ref{single-scale-comparison}.}
\label{head}
\end{figure}

To enhance the representation ability of feature pyramids, we propose a simple yet effective method, called as receptive field pyramids (RFP). Our main motivation is to let the network have different receptive fields adaptively in each feature map based on the varying scales of detected faces. To this end,  we design a module of multiple parallel branches with various receptive fields.  We let each branch within this module shares the same structure and weights yet has different receptive fields using dilated convolutions\cite{Yu2016}.  Then we use a branch pooling  to fuse information from different parallel branches. The branch pooling balances the representations of parallel branches during training and enables a single branch to implement inference during testing, which further reduces the inference cost significantly. Empirically,  the proposed method can accelerate the inference speed significantly while achieving state-of-the-art performance, as  Figure \ref{head} shown.

In a nutshell, our main contributions in the present work can be summarized as follows:

(1) We propose a receptive field pyramids method to enhance the representations of feature pyramids, where multiple parallel branches are designed with the same convolutional weights yet different dilation rates.

(2) We design a branch pooling operator to balance the representations of parallel branches during training and enables a single branch to implement inference during testing, which further accelerates the inference speed.

(3) We verify the superiority of our proposed method on two face detection datasets---WIDER FACE and UFDD. Our method can achieve state-of-the-art performances with a notably fast inference speed.

\begin{figure*}[htb]
\centering
\includegraphics[width=\linewidth]{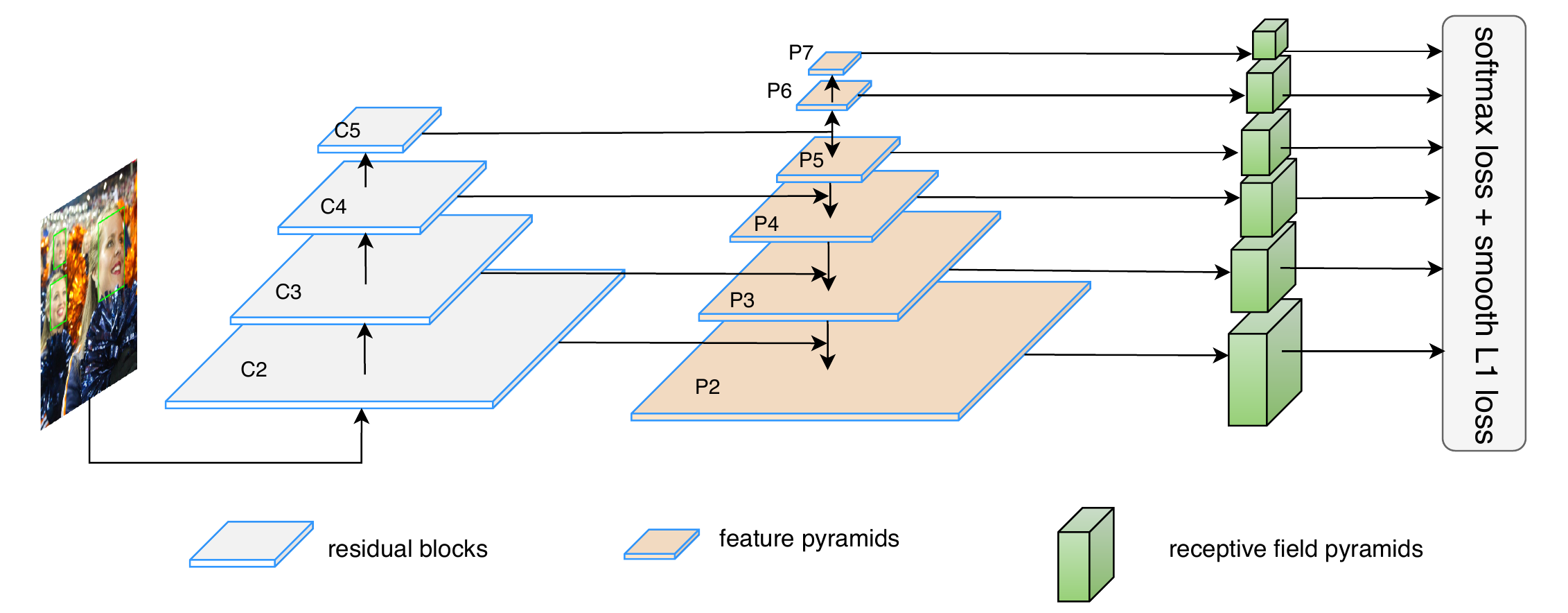}
\caption{The architecture of the proposed face detector: EMface, including three parts: backbone, feature pyramid network(FPN), and receptive field pyramids (RFP). The backbone network extracts features from the raw input image and generates four-level feature maps C2-C5. FPN enhances the feature maps from the backbone and creates six-level feature pyramids. Each-level feature map is appended as an independent RFP module. Following RFP, a multi-task loss is used for end-to-end optimization.}
\label{structure}
\end{figure*}

\section{Related Work}
The CNN-based feature  have gradually replaced the traditional hand-crafted feature extractors in face detection. Li et al.\cite{Li2015} proposed a cascade architecture built on CNNs for face detection. The detector rejects false detections in the early stages and verifies the detections in the later stages. MTCNN\cite{ZhangK2016} proposed a cascaded structure with three stages of CNNs by multi-task learning for joint face detection and alignment. S\textsuperscript{3}FD\cite{Zhang_ICCV2017} was one of the first to inherit the framework of SSD\cite{Liu2016} for face detection. The basic structure of S\textsuperscript{3}FD was employed by PyramidBox\cite{Tang_2018_ECCV} and DSFD\cite{DSFD2019} associated with Feature Pyramids Network (FPN)\cite{Lin2017} to improve semantics in low-level feature maps. SRN\cite{Chi2019} introduced two-step classification and regression operations selectively into an anchor-based face detector to reduce false positives and improve location accuracy simultaneously.

Many studies that discuss receptive fields for recognition tasks in computer vision. In ASPP\cite{Deeplab} that was proposed for semantic segmentation, feature representation was generated by concatenating atrous-convolved features with different dilation rates. RFB Net\cite{Liu_2018_ECCV} proposed a block by combining multiple branches with different kernels and dilated convolution layers to enhance the feature discriminability. Deformable convolution\cite{dai17dcn} attempted to adaptively adjust the spatial distribution of receptive fields according to the scale and shape of the object, which added 2D offsets to the regular grid sampling locations in the standard convolution. 

\section{Our Method}
\subsection{Revisiting Face Detection Architecture}
Most of the modern face detectors inherit the framework of SSD network \cite{liu2016ssd}, where VGG \cite{Simonyan14} or ResNet\cite{He2016} is used as its backbone network. For detecting small faces, the feature pyramids network  (FPN) is employed by recovering high-resolution representations from low-resolution ones via a top-down pathway and lateral connections.

As shown in Figure \ref{structure}, we adopt ResNet and FPN to construct a 6-level feature pyramids structure as the backbone network. The feature maps extracted from four stages' last residual blocks of ResNet are denoted as \big\{C2, C3, C4, C5\big\}, respectively. They have strides of \big\{4, 8, 16, 32\big\} pixels with respect to the input image. We use top-down and lateral connections to generate feature maps---\big\{P2, P3, P4, P5\big\}. In addition, P6 and P7 are down-sampled by two $3 \times 3$ convolution layers with a stride of 2 pixels after P5. The final set of feature pyramids is called \big\{P2, P3, P4, P5, P6, P7\big\}, where each feature map has 256-channel outputs. We append a receptive field pyramids module after each feature map in feature pyramids. A multi-task loss is used for end-to-end optimization, where softmax loss and smooth L1 loss are for classification and regression, respectively.
\begin{figure}[htb]
\centering
\includegraphics[width=\linewidth]{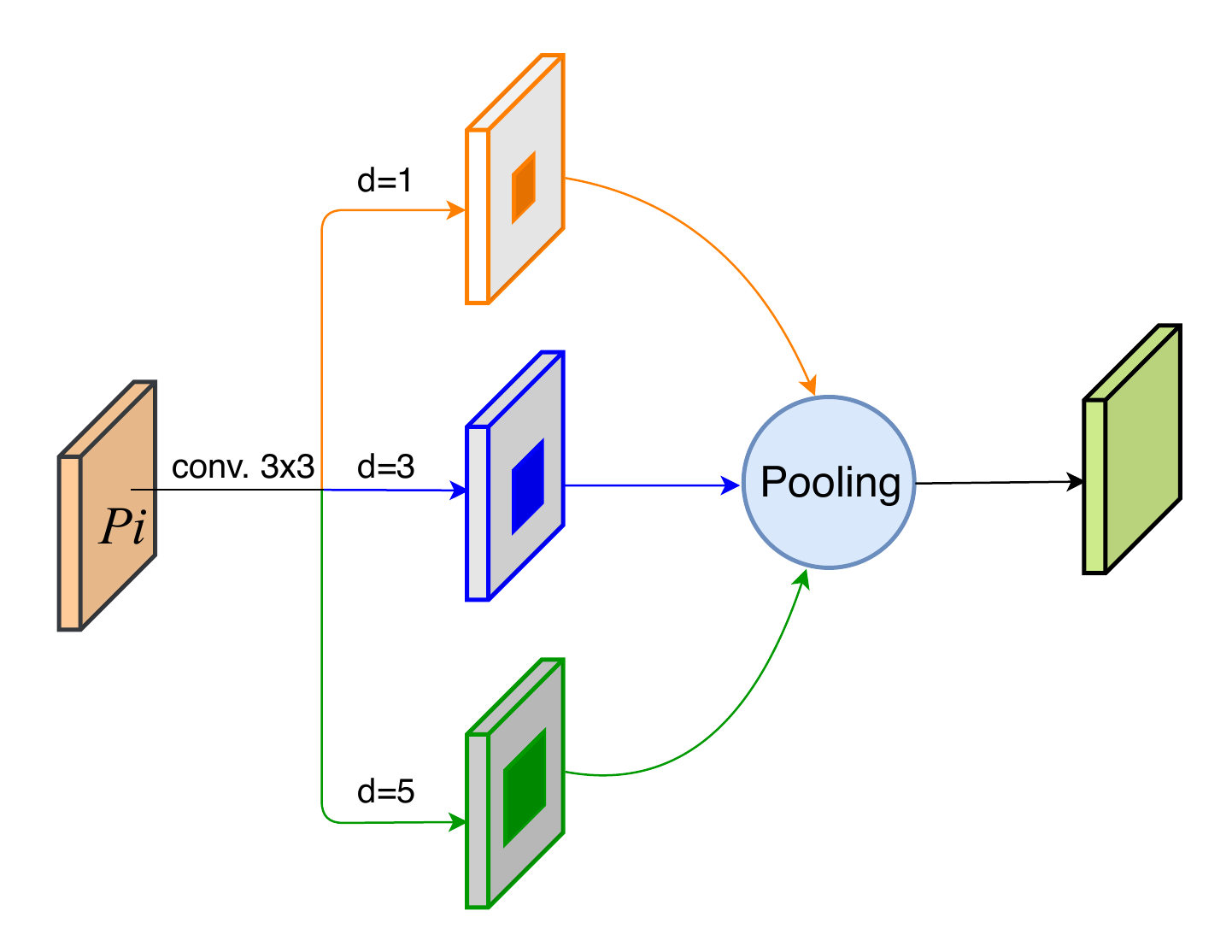}
\caption{The structure of the receptive field pyramids module.}
\label{RFP}
	\vspace{-0.3cm} 
\end{figure}
\subsection{Receptive Field Pyramids}
As  Figure \ref{RFP}  shown, the receptive field pyramids (RFP) can be divided into two main components: the multi-branch convolution layer and the branch pooling layer. The former part is to provide the input feature map different receptive fields, and the latter one is to fuse the receptive field pyramids for the target face. The output still has the same size as the input feature map, including the resolution and the number of channels. 

\subsubsection{Multiple Branches.} 
The module consists of three parallel branches of dilated convolutions with the same kernel size but with different dilation rates. To be specific, we use a dilated 3$\times$3 convolution with dilation rate 1, 3, and 5 for different branches, as shown in Figure \ref{RFP}. We also apply the shortcut design from ResNet\cite{He2016} for each branch. Formally, in this paper, we consider an RFP module defined as:
\begin{equation}
\begin{split}
 & y_{1}=W_{1}^{d=1}x+x \\
 & y_{2}=W_{2}^{d=3}x+x \\
 & y_{3}=W_{3}^{d=5}x+x
\end{split}
\end{equation}
Here $x$ and $ y_{i}(i=1,2,3) $ are the input and output vectors of the branch considered. And $ W_{1}^{d=1}, W_{2}^{d=3}, W_{3}^{d=5} $ denote the convolutional weights in three branches, where $d$ denotes the dilation rate. To further decrease the number of parameters, we let different branches share the same weights and only vary the dilation rate of each branch: $ W_{1}=W_{2}=W_{3} $.
\subsubsection{Branch Pooling.}
The outputs of different parallel branches are generally concatenated together following a 1$\times$1 convolution layer that is to decrease the number of channels, as seen in ASPP\cite{Deeplab}, RFB\cite{Liu_2018_ECCV}, etc. To avoid introducing additional parameters in our module, we thus propose \textit{Branch Pooling} to fuse information from different parallel branches. Suppose the outputs of multiple branches are: $y_{1}\in R^{H\times W\times C}$, $y_{2}\in R^{H\times W\times C}$,$\cdots$, $y_{B}\in R^{H\times W\times C}$, where H and W are the spatial height and width axes, C is the channel axis, and B is the number of branches. We compute the average-pooled features along the B axis. In this paper, we let $B=3$:
\begin{equation}
y_{pooling}=\frac{1}{B}\sum_{i=1}^{B}y_{i}
\end{equation}
The averaging operation can balance the representations of different parallel branches during training, which enables a single branch to implement inference during testing.
\section{Experiments}
In this section, we first conduct the ablation studies on receptive field pyramids. We then report the comparison results of our face detector, named EMface, with state-of-the-art methods on the WIDER FACE\cite{Yang2016} validation set and the UFDD dataset\cite{Nada2018}. The metric to evaluate the detection performance is the average precision (AP) at the IoU threshold 0.5.
\begin{table}[htbp]
\small
\centering
\caption{Single-scale testing results of EMface that uses different number of branches in receptive field pyramids (RFP) on the WIDER FACE validation set. The GFLOPs is calculated on the input size 960$\times$1024.}
\label{Branches}
\begin{tabular}{c|c|c|ccc}
\hline
Branches & Params. & GFLOPs & Easy & Medium & Hard \\
\hline
\hline
Baseline & 26.19M & 132.91 & 94.6 & 93.6 & 85.2 \\
\hline
1 & 29.56M & 178.02 & 94.6 & 93.7 & 85.5 \\
2 & 29.56M & 223.11 & 94.8 & 93.8 & 87.0 \\
3 & 29.56M & 268.20 & 95.0 & 94.0 & 87.2 \\
4 & 29.56M & 313.29 & 95.1 & 94.1 & 87.2 \\
\hline
\end{tabular}
\end{table}
\subsection{Ablation Study}
In this subsection, we empirically show the effectiveness of our design choice in the RFP module. We firstly construct a simple baseline based on the architecture, as shown in Figure \ref{structure}, where ResNet50 associated with FPN is adopted as the backbone network yet without appending RFP modules to feature pyramids. We train all models in the training set of WIDER FACE and evaluate on the validation set. 
\begin{table}[htbp]
\centering
\caption{Single-scale testing results of EMface that uses different weight sharing techniques on the WIDER FACE validation set. The GFLOPs is calculated on the input size 960$\times$1024.}
\label{Weight-sharing}
\small
\begin{tabular}{c|c|c|ccc}
\hline
Method & Params. & GFLOPs & Easy & Medium & Hard \\
\hline
\hline
Baseline & 26.19M & 132.91 & 94.6 & 93.6 & 85.2 \\
\hline
W/o sharing & 36.33M & 268.20 & 95.3 & 94.3 & 87.3 \\
W sharing & 29.56M & 268.20 & 95.0 & 94.0 & 87.2 \\
\hline
\end{tabular}
\end{table}
\subsubsection{Number of Branches.}
In this part, we investigate whether the number of branches in each module would affect the performance of EMface. Table \ref{Branches} shows the results of EMface with one to four branches in each module. When there is one branch in the module, we use a conventional 3$\times$3 convolution layer. It shows that it can not significantly improve the performance of the baseline. EMface with two branches (d=1 and d=3) in each module have been obviously improved on the hard subset, which is further improved when there are three branches in each module. We observe that four branches can not bring further significant improvement over three branches. Thus, three branches in each module are used considering a tradeoff between computational cost and performance of EMface.
\begin{table}[htbp]
\small
\centering
\caption{Single-scale testing results on the WIDER FACE validation set of EMface with single branch inference in RFP where branch pooling (BP), adding operator and concatenating operator are respectively used when training. The GFLOPs is calculated on the input size 960$\times$1024.}
\label{Single_branch_AP}
\begin{tabular}{c|c|c|ccc}
\hline
Method & Branch No. & GFLOPs & Easy & Medium & Hard \\
\hline
\hline
Baseline &  - & 132.91 & 94.6 & 93.6 & 85.2 \\
\hline
\multirow{4}{*}{BP} & Branch-1 & 178.02 & 94.6 & 93.8 & 87.2 \\
 ~ & Branch-2 & 178.02 & 95.1 & 94.0 & 87.2 \\
 ~ & Branch-3 & 178.02 & 95.1 & 93.9 & 87.2 \\
 ~ & 3 Branches & 268.20 & 95.0 & 94.0 & 87.2 \\
\hline
\multirow{4}{*}{Add} & Branch-1 & 178.02 & 93.7 & 92.7 & 80.5 \\
 ~ & Branch-2 & 178.02 & 94.5 & 92.3 & 80.2 \\
 ~ & Branch-3 & 178.02 & 94.3 & 92.1 & 78.9 \\
 ~ & 3 Branches & 268.20 & 94.8 & 93.6 & 87.1 \\
\hline
\multirow{4}{*}{Cat} & Branch-1 & 178.02 & N/A & N/A & N/A \\
 ~ & Branch-2 & 178.02 & N/A & N/A & N/A \\
 ~ & Branch-3 & 178.02 & N/A & N/A & N/A \\
 ~ & 3 Branches & 298.19 & 95.1 & 94.2 & 87.1 \\
\hline
\end{tabular}
\end{table}
\begin{table*}[htbp]
\centering
\caption{Comparisons on the state-of-the-art face detectors on WIDER FACE validation set with the \textbf{\textit{single-scale}} testing.}
\label{single-scale-comparison}
\begin{spacing}{1.2}
\begin{tabular}{c|c|c|ccc|c}
\hline
Method & Backbone & Params. & Easy & Medium & Hard & Runtime(ms/image) \\
\hline
\hline
 S\textsuperscript{3}FD & VGG16 & 21.42M & 93.7 & 92.4 & 83.4 & 466 \\
 PyramidBox & VGG16+FPN & 54.53M  & 94.8 & 93.9 & 86.5 & 244 \\
 SRN & ResNet50+FPN & 50.90M & 95.2 & 93.1 & 86.4 & 256 \\
 DSFD & ResNet152+FPN & 114.50M  & 94.8  & 93.4 & 84.6 & 508 \\
\hline
\hline
Baseline & ResNet50+FPN & 26.19M  & 94.6 & 93.6 & 85.2  & 177 \\
\hline
 Ours & VGG16+FPN & 26.31M  & 94.8 & 94.0 & 87.8 & 186 \\
 Ours & ResNet50+FPN & 29.56M & 95.1 & 94.0 & 87.2 & 182 \\
\hline
\end{tabular}
\end{spacing}
\end{table*}
\subsubsection{Weight Sharing.}
Table \ref{Weight-sharing} shows the effectiveness of applying weight sharing on multi-branch. In the table, `W/o sharing' denotes different branches in each RFP module that do not share weights. `W sharing' denotes different branches in each RFP module share the same weights, which is our default configuration. Different branches in each module have independent weights that can make EMface acquire better performance, especially on Easy and Medium subsets, yet with more additional parameters. Weight sharing in each module only slightly decreases the detection precision of EMface comparing with the result of `W/o sharing', yet reduces a large number of parameters. 
\subsubsection{Branch Pooling.}
In this part, we compare our proposed branch pooling (BP) operator, with two commonly used operators---adding and concatenating, when fusing information from different feature maps. Experimental results with various aggregation methods in the RFP module are shown in Table \ref{Single_branch_AP}. Since each RFP has three branches to represent features, we receptively evaluate EMface with a single branch and three branches inference in each RFP. For the single branch inference, we receptively drop out the branch pooling, adding or concatenating operators, and output a single branch's results. We observe that these three methods make EMface acquire similar performance when EMface is with three branches inference. Note that BP and adding operators have the same computational costs, yet the concatenating operator requires additional parameters and computational costs. 

As expected, the BP operator balances each branch's representational power, which achieves the same results on the hard subset. However, the adding operator disables single branch inference to make as good results as three branches. The concatenating operator does not produce valid predictions when using single branch inference. The single branch inference of BP achieves the same results with three branches inference, and it can further reduce the inference time. Thus, we drop out the Branch-1 and Branch-3 (d=1 and d=5 in Figure \ref{RFP}) in RFP and only keep the Branch-2 to output in the inference phase. This experiment demonstrates the effectiveness and superiority of our proposed branch pooling, which enables RFP to achieve a single branch inference while keeping as good performance as three branches inference.
\begin{figure}[htbp]
	\centering
	\includegraphics[width=\linewidth]{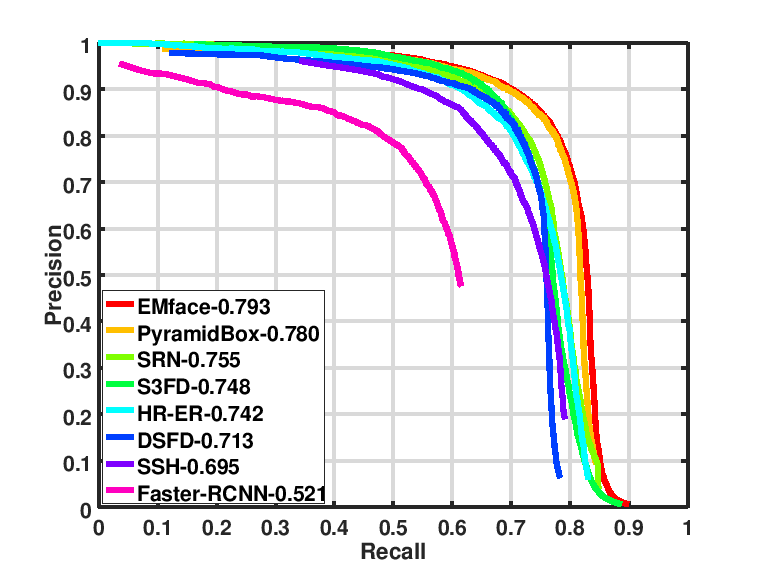}
	\caption{Precision-recall curves on the UFDD dataset.}
	\label{UFDD}
\end{figure}
\subsection{Comparison with State-of-the-Arts}
\subsubsection{WIDER FACE.}
In most of the previous works, they only report the multi-scale testing results on the WIDER FACE benchmark. It is difficult to fairly compare different face detectors on inference speed versus detection precision, since different methods may differ in multi-scale testing strategies. Thus, we evaluate EMface and four state-of-the-art methods, i.e., S\textsuperscript{3}FD\cite{Zhang_ICCV2017}, PyramidBox\cite{Tang_2018_ECCV}, SRN\cite{Chi2019}, and DSFD\cite{DSFD2019}, on the single-scale testing in terms of accuracy and speed. Detectors run on a single V100 GPU with CUDA 9 and CUDNN 7 with a batch size of 1. Note that we do not use CUDA accelerating NMS operation. Results are reported in Table \ref{single-scale-comparison}. We choose two different networks as our backbone, including VGG16+FPN and ResNet50+FPN. Compared with other methods, our face detector achieves much better results on the hard subset, with less number of parameters and runtime cost. 

\subsubsection{UFDD.}
To demonstrate the generalization performance of EMface, we evaluate it and compare it with several state-of-the-art methods on the UFDD benchmark. The single-scale testing results are shown in Figure \ref{UFDD}, which illustrates that EMface surpasses all other methods. The superiority of EMface demonstrates that EMface is robust to the weather-based degradations, blur, and other unconstrained conditions.
\section{Conclusion}
In this paper, we proposed a receptive field pyramids method for strengthening the representational power of feature pyramids on faces of different scales in face detection. The receptive filed pyramids could diversify the receptive fields of each detecting face, using multiple parallel branches with the same convolutional weights yet different dilation rates. We then developed a branch pooling operator to balance the representations of parallel branches during training and enables a single branch to implement inference during testing, which further reduces the inference time. We evaluated the performance of our method with application to face detection on two benchmark datasets. The empirical results verified that the proposed method could achieve state-of-the-art performance with a fast inference speed. Future work will involve facilitating such a trend and applying this receptive field pyramids method to more widespread applications, such as general object detection, and semantic segmentation.
\bibliographystyle{IEEEbib}
\bibliography{icme2020template}
\end{document}